\documentclass{article}
\usepackage{spconf,amsmath,amssymb,graphicx}
\usepackage{booktabs}
\usepackage{url}

\title{Federated learning for mobile keyboard prediction}

\name{Andrew Hard, Kanishka Rao, Rajiv Mathews, Swaroop Ramaswamy, Fran{\c{c}}oise Beaufays}
\secondname{Sean Augenstein, Hubert Eichner, Chlo{\'{e}} Kiddon, Daniel Ramage}

\address{
  Google LLC,\\
  Mountain View, CA, U.S.A.\\
  \texttt{\{harda, kanishkarao, mathews, swaroopram, fsb}\\
  \texttt{saugenst, huberte, loeki, dramage\}@google.com}
}

\begin{document}
\maketitle

\begin{abstract}

We train a recurrent neural network language model using a distributed,
on-device learning framework called federated learning for the purpose of
next-word prediction in a virtual keyboard for smartphones. Server-based
training using stochastic gradient descent is compared with training on client
devices using the \texttt{FederatedAveraging} algorithm. The federated
algorithm, which enables training on a higher-quality dataset for this use case,
is shown to achieve better prediction recall. This work demonstrates the
feasibility and benefit of training language models on client devices without
exporting sensitive user data to servers. The federated learning environment
gives users greater control over the use of their data and simplifies the task
of incorporating privacy by default with distributed training and aggregation
across a population of client devices.

\end{abstract}


\section{Introduction}
\label{sec:introduction}

Gboard --- the Google keyboard\footnotemark --- is a virtual keyboard for
touchscreen mobile devices with support for more than 600 language varieties and
over 1 billion installs as of 2019. In addition to decoding noisy signals from
input modalities including tap and word-gesture typing, Gboard provides
auto-correction, word completion, and next-word prediction features.

\footnotetext{gboard.app.goo.gl/get}

\begin{figure}
  \centering
  \includegraphics[width=0.85\columnwidth]{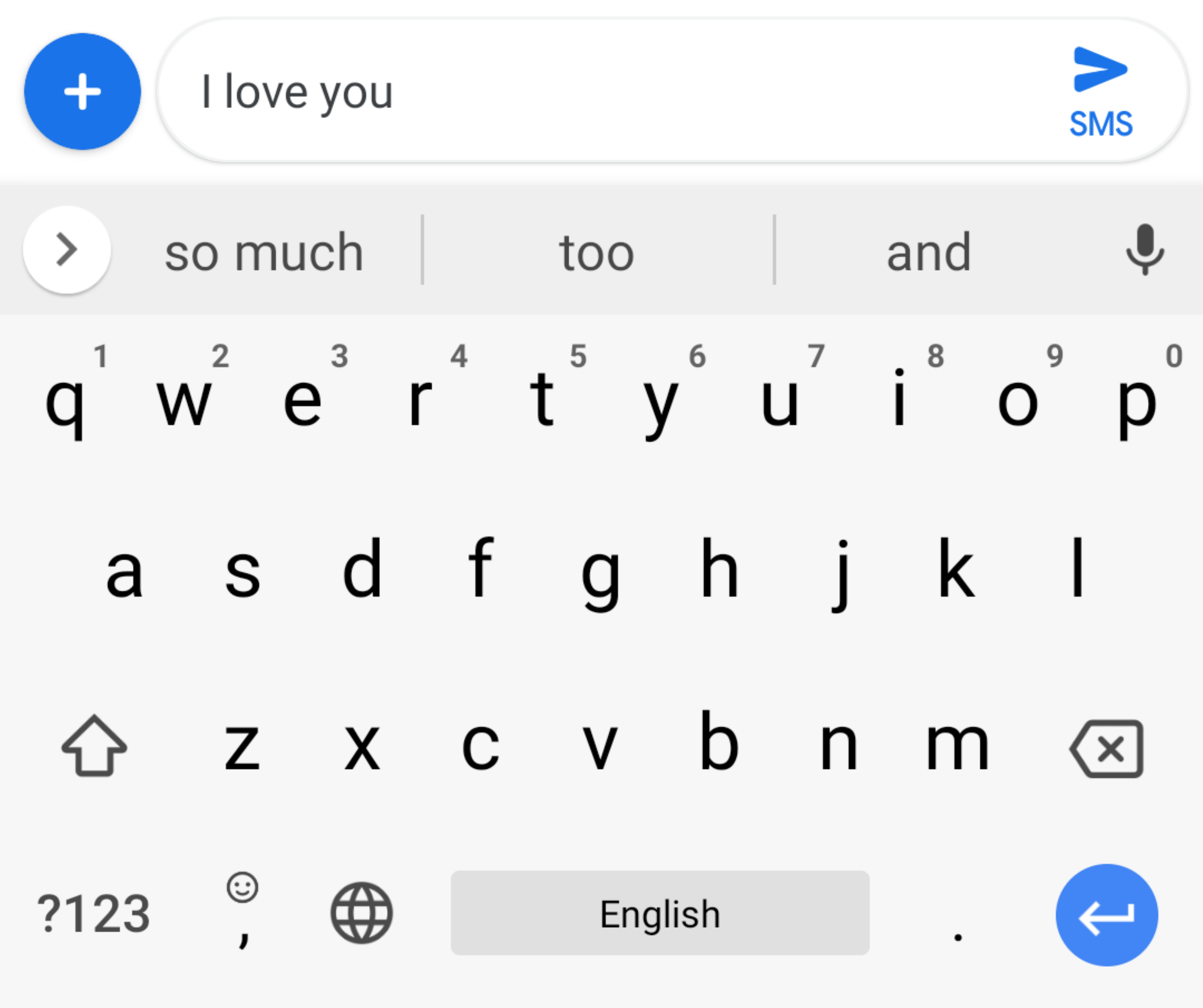}
  \caption{Next word predictions in Gboard. Based on the context ``I love you'',
  the keyboard predicts ``and'', ``too'', and ``so much''.}
  \label{fig:gboard_screenshot}
\end{figure}

As users increasingly shift to mobile devices~\cite{pewinternet}, reliable and
fast mobile input methods become more important. Next-word predictions provide a
tool for facilitating text entry. Based on a small amount of user-generated
preceding text, language models (LMs) can predict the most probable next word or
phrase. Figure~\ref{fig:gboard_screenshot} provides an example: given the text,
``I love you'', Gboard predicts the user is likely to type ``and'', ``too'', or
``so much'' next. The center position in the suggestion strip is reserved for
the highest-probability candidate, while the second and third most likely
candidates occupy the left and right positions, respectively.

Prior to this work, predictions were generated with a word n-gram finite state
transducer (FST)~\cite{fst_nlp}. The mechanics of the FST decoder in Gboard ---
including the role of the FST in literal decoding, corrections, and completions
--- are described in Ref.~\cite{gboard_fst}. Next word predictions are built by
searching for the highest-order n-gram state that matches the preceding text.
The $n$-best output labels from this state are returned. Paths containing
back-off transitions to lower-orders are also considered. The primary (static)
language model for the English language in Gboard is a Katz smoothed Bayesian
interpolated~\cite{bayeslm} 5-gram LM containing 1.25 million n-grams, including
164,000 unigrams. Personalized user history, contacts, and email n-gram models
augment the primary LM.

Mobile keyboard models are constrained in multiple ways. In order to run on both
low and high-end devices, models should be small and inference-time latency
should be low. Users typically expect a visible keyboard response within 20
milliseconds of an input event. Given the frequency with which mobile keyboard
apps are used, client device batteries could be quickly depleted if CPU
consumption were not constrained. As a result, language models are usually
limited to tens of megabytes in size with vocabularies of hundreds of thousands
of words.

Neural models --- in particular word and character-level recurrent neural
networks (RNNs)~\cite{rnn} --- have been shown to perform well on language
modeling tasks ~\cite{lstm, neuralproblm, charawarenlm}. Unlike n-gram models
and feed-forward neural networks that rely on a fixed historical context window,
RNNs utilize an arbitrary and dynamically-sized context window. Exploding and
vanishing gradients in the back-propagation through time algorithm can be
resolved with the Long Short-Term Memory (LSTM)~\cite{lstm}. As of writing,
state-of-the art perplexities on the 1 billion word benchmark~\cite{onebillion}
have been achieved with LSTM variants~\cite{limitsoflm, outrageouslm}.

Training a prediction model requires a large data sample that is representative
of the text that users will commit. Publicly available datasets can be used,
though the training distribution often does not match the population's
distribution. Another option is to sample user-generated text. This requires
logging, infrastructure, dedicated storage on a server, and security. Even with
data cleaning protocols and strict access controls, users might be uncomfortable
with the collection and remote storage of their personal data~\cite{fedlearn}.

In this paper, we show that federated learning provides an alternative to the
server-based data collection and training paradigm in a commercial setting. We
train an RNN model from scratch in the server and federated environments and
achieve recall improvements with respect to the FST decoder baseline.

The paper is organized in the following manner. Section~\ref{sec:related_work}
summarizes prior work related to mobile input decoding, language modeling with
RNNs, and federated learning. Coupled Input-Forget Gates (CIFG) --- the RNN
variant utilized for next-word prediction --- are described in
Section~\ref{sec:model}. Section~\ref{sec:federated} discusses the federated
averaging algorithm in more depth. Section~\ref{sec:experiments} summarizes
experiments with federated and server-based training of the models. The results
of the studies are presented in Section~\ref{sec:results}, followed by
concluding remarks in Section~\ref{sec:conclusion}.

\section{Related Work}
\label{sec:related_work}

FSTs have been explored in the context of mobile keyboard input decoding,
correction, and prediction~\cite{gboard_fst}. LSTMs have greatly improved the
decoding of gestured inputs on mobile keyboards~\cite{nsm}. RNN language models
optimized for word prediction rate and keystroke savings within inference-time
latency and memory constraints have also been published~\cite{rnnlm, kbdnn}.

Research into distributed training for neural models has gained relevance with
the recent increased focus on privacy and government regulation. In particular,
federated learning has proved to be a useful extension of server-based
distributed training to client device-based training using locally stored
data~\cite{fedlearn, privatedl}. Language models have been trained using the
federated algorithm combined with differential
privacy~\cite{differential-privacy, dplm}. And Gboard has previously used
federated learning to train a model to suggest search queries based on typing
context~\cite{fedblog}, though the results have not been published yet. To the
best of our knowledge, there are no existing publications that train a neural
language model for a mobile keyboard with federated learning.

\section{Model Architecture}
\label{sec:model}

The next-word prediction model uses a variant of the Long Short-Term Memory
(LSTM)~\cite{lstm} recurrent neural network called the Coupled Input and Forget
Gate (CIFG)~\cite{cifg}. As with Gated Recurrent Units~\cite{GRU}, the CIFG uses
a single gate to control both the input and recurrent cell self-connections,
reducing the number of parameters per cell by 25\%. For timestep $t$, the input
gate ${i}_{t}$ and forget gate ${f}_{t}$ have the relation:

\begin{equation}
  {f}_{t} = 1 - {i}_{t}.
\end{equation}

The CIFG architecture is advantageous for the mobile device environment because
the number of computations and the parameter set size are reduced with no impact
on model performance. The model is trained using TensorFlow~\cite{tensorflow}
without peephole connections. On-device inference is supported by TensorFlow
Lite\footnotemark.

\footnotetext{https://www.tensorflow.org/lite/}

Tied input embedding and output projection matrices are used to reduce the model
size and accelerate training~\cite{shareioemb,tyingioemb}. Given a vocabulary of
size $V$, a one-hot encoding $v \in {\mathbb{R}}^{V}$ is mapped to a dense
embedding vector $d \in {\mathbb{R}}^{D}$ by $d = W v$ with an embedding matrix
$W \in {\mathbb{R}}^{D \times V}$. The output projection of the CIFG, also in
${\mathbb{R}}^{D}$, is mapped to the output vector ${W}^{\mathsf{T}} h \in
{\mathbb{R}}^{V}$. A softmax function over the output vector converts the raw
logits into normalized probabilities. Cross-entropy loss over the output and
target labels is used for training.

The client device requirements alluded to in Section~\ref{sec:introduction}
limit the vocabulary and model sizes. A dictionary of $V=\text{10,000}$ words is
used for the input and output vocabularies. Input tokens include special
beginning of sentence, end of sentence, and out-of-vocabulary tokens. During
network evaluation and inference, the logits corresponding to these special
tokens are ignored. The input embedding and CIFG output projection dimension $D$
is set to 96. A single layer CIFG with 670 units is used. Overall, 1.4 million
parameters comprise the network --- more than two thirds of which are associated
with the embedding matrix $W$. After weight quantization, the model shipped to
Gboard devices is 1.4 megabytes in size.

\section{Federated Learning}
\label{sec:federated}

\begin{figure}
  \centering
  \includegraphics[width=0.99\columnwidth]{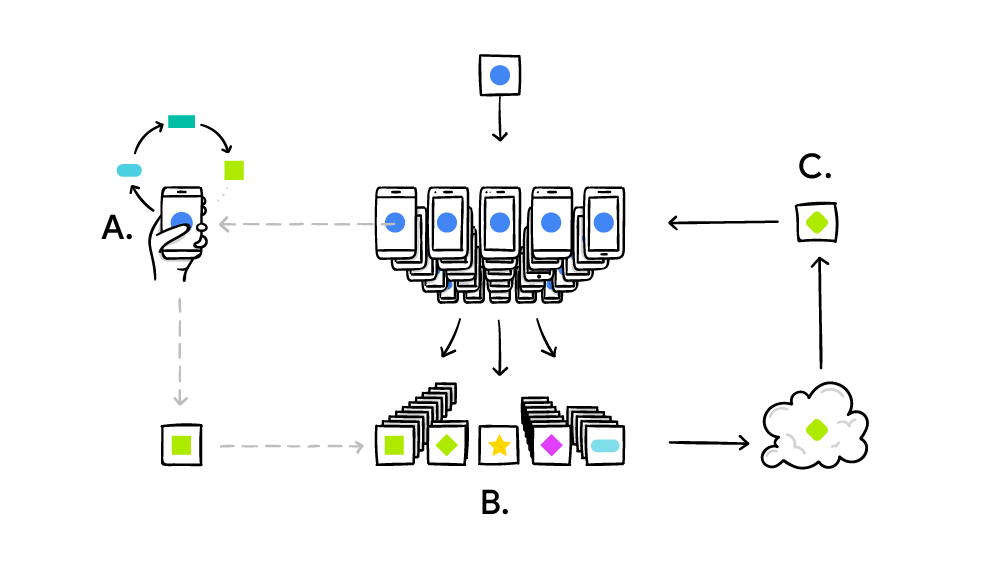}
  \caption{An illustration of the federated learning process from
           Ref.~\cite{fedblog}: (A) client devices compute SGD updates on
           locally-stored data, (B) a server aggregates the client updates
           to build a new global model, (C) the new model is sent back to
           clients, and the process is repeated.}
  \label{fig:illustration_fed}
\end{figure}

Federated learning~\cite{fedlearn, privatedl} provides a decentralized
computation strategy that can be employed to train a neural model. Mobile
devices, referred to as clients, generate large volumes of personal data that
can be used for training. Instead of uploading data to servers for centralized
training, clients process their local data and share model updates with the
server. Weights from a large population of clients are aggregated by the server
and combined to create an improved global model.
Figure~\ref{fig:illustration_fed} provides an illustration of the process. The
distributed approach has been shown to work with unbalanced datasets and data
that are not independent or identically distributed across clients.

The \texttt{FederatedAveraging} algorithm~\cite{fedlearn} is used on the server
to combine client updates and produce a new global model. At training round $t$,
a global model ${w}_{t}$ is sent to a subset $K$ of client devices. In the
special case of $t=0$, client devices start from the same global model that has
either been randomly initialized or pre-trained on proxy data. Each of the
clients participating in a given round has a local dataset consisting of
${n}_{k}$ examples, where $k$ is an index of participating clients. ${n}_{k}$
varies from device to device. For studies in Gboard, ${n}_{k}$ is related to the
user's typing volume.

Every client computes the average gradient, ${g}_{k}$, on its local data with
the current model ${w}_{t}$ using one or more steps of stochastic gradient
descent (SGD). For a client learning rate $\epsilon$, the local client update,
${w}_{t+1}^{k}$, is given by:

\begin{equation}
  {w}_{t} - \epsilon {g}_{k} \rightarrow {w}_{t+1}^{k}.
\end{equation}

The server then does a weighted aggregation of the client models to obtain a new
global model, ${w}_{t+1}$:

\begin{equation}
  \sum_{k=1}^{K} \frac{{n}_{k}}{N} {w}_{t+1}^{k} \rightarrow {w}_{t+1},
  \label{eq:server_weight}
\end{equation}

\noindent
where $N = \sum_{k} {n}_{k}$. In essence, the clients compute SGD updates
locally, which are communicated to the server and aggregated. Hyperparameters
including the client batch size, the number of client epochs, and the number of
clients per round (global batch size) are tuned to improve performance.

Decentralized on-device computation offers fewer security and privacy risks than
server storage, even when the server-hosted data are anonymized. Keeping
personal data on client devices gives users more direct and physical control of
their own data. The model updates communicated to the server by each client are
ephemeral, focused, and aggregated. Client updates are never stored on the
server; updates are processed in memory and are immediately discarded after
accumulation in a weight vector. Following the principle of data
minimization~\cite{privacywh}, uploaded content is limited to model weights.
Finally, the results are only used in aggregate: the global model is improved
by combining updates from many client devices. The federated learning procedure
discussed here requires users to trust that the aggregation server will not
scrutinize individual weight uploads. This is still preferable to server
training because the server is never entrusted with user data. Additional
techniques are being explored to relax the trust requirement. Federated learning
has previously been shown to be complementary to privacy-preserving techniques
such as secure aggregation~\cite{secagg} and differential
privacy~\cite{differential-privacy}.

\section{Experiments}
\label{sec:experiments}

Federated learning and server-based stochastic gradient descent are used to
train the CIFG language model described in Section~\ref{sec:model} starting from
random weight initializations. The performance of both models is evaluated on
server-hosted logs data, client-held data, and in live production experiments.

\subsection{Server-based training with logs data} \label{sec:server_training}

\begin{figure}
  \centering
  \includegraphics[width=0.95\columnwidth]{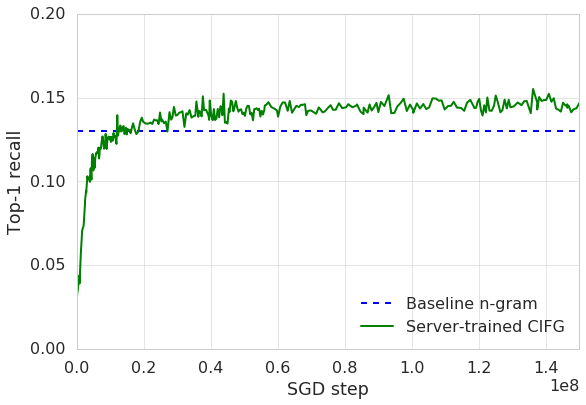}
  \caption{Top-1 recall of the CIFG as a function of SGD step during server
           training. The recall of the n-gram FST baseline model is shown for
           comparison, but the FST model is not trained in this study.}
  \label{fig:server_accuracy}
\end{figure}

Server-based training of the CIFG next-word prediction model relies on data
logged from Gboard users who have opted to share snippets of text while typing
in Google apps. The text is truncated to contain short phrases of a few words,
and snippets are only sporadically logged from individual users. Prior to
training, logs are anonymized and stripped of personally identifiable
information. Additionally, snippets are only used for training if they begin
with a start of sentence token.

For this study, logs are collected from the English speaking population of
Gboard users in the United States. Approximately 7.5 billion sentences are used
for training, while the test and evaluation samples each contain 25,000
sentences. The average sentence length in the dataset is 4.1 words. A breakdown
of the logs data by app type is provided in Table~\ref{tab:app_logs}. Chat apps
generate the majority of logged text.

Asynchronous stochastic gradient descent with a learning rate equal to
${10}^{-3}$ and no weight decay or momentum is used to train the server CIFG.
Adaptive gradient methods including Adam~\cite{adam} and AdaGrad~\cite{adagrad}
are not found to improve the convergence. Sentences are processed in batches of
50. The network converges after 150 million steps of SGD.
Figure~\ref{fig:server_accuracy} shows the top-1 recall of the CIFG during
network training, compared with the performance of the n-gram baseline model.

\begin{table}[!h]
  \centering
  \begin{tabular}{lc} \toprule
    App type       & Share of data \\ \hline \midrule
    Chat           & 60\% \\
    Web input      & 35\% \\
    Long form text & 5\%  \\
  \bottomrule
  \end{tabular}
  \caption{The composition of logs data by mobile app type.}
  \label{tab:app_logs}
\end{table}

\subsection{Federated training with client caches} \label{sec:client_cache}

\begin{figure}
  \centering
  \includegraphics[width=0.95\columnwidth]{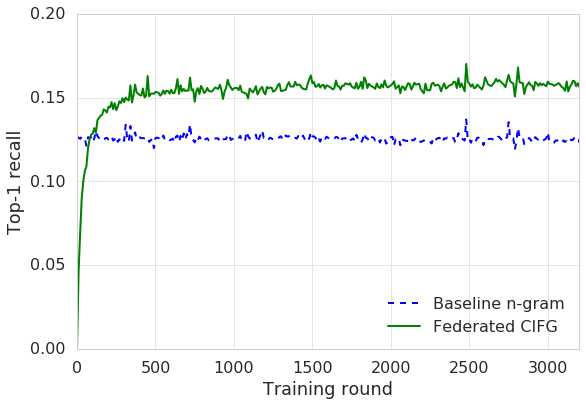}
  \caption{Top-1 recall of the CIFG as a function of training round during
           federated training. The performance of the n-gram FST baseline model
           is evaluated on the client caches along with the CIFG, but it is not
           trained in this study.}
  \label{fig:accuracy_fed}
\end{figure}

Data for the federated training of the CIFG next-word prediction model are
stored in local caches on Gboard client devices. As with the logs data, each
client cache stores text belonging to the device owner, as well as prediction
candidates generated by the decoder.

Client devices must meet a number of requirements in order to be eligible for
federated training participation. In terms of hardware requirements, the devices
must have at least 2 gigabytes of memory available. Additionally, the clients
are only allowed to participate if they are charging, connected to an un-metered
network, and idle. These criteria are chosen specifically for the Gboard
implementation of federated learning and are not inherent to the federated
learning platform. Clients for this study are also required to be located in
North America while running Gboard release 7.3 or greater with the US English
language model enabled.

Unlike server-based training, where train, test, and eval samples are obtained
via explicit splits of the data, the federated train, test, and eval samples are
obtained by defining separate computation tasks. While there is no explicit
separation of client devices into three distinct populations, the probability of
client reuse in both the training and test or eval tasks is minimal in a
sufficiently large client population. The composition of the client cache data
by app type is shown in Table~\ref{tab:app_fed}. As with the logs data, the
client caches are also dominated by chat apps. Social media apps have an
increased presence in the client cache sample, while long-form communication is
represented less.

\begin{table}[!h]
  \centering
  \begin{tabular}{lc} \toprule
    App type  & Share of data \\ \hline \midrule
    Chat      & 66\% \\
    Social    & 16\% \\
    Web input & 5\%  \\
    Other     & 12\% \\
  \bottomrule
  \end{tabular}
  \caption{The composition of client cache data by mobile app type.}
  \label{tab:app_fed}
\end{table}

The \texttt{FederatedAveraging} algorithm described in
Section~\ref{sec:federated} is used to aggregate distributed client SGD updates.
Between 100 and 500 client updates are required to close each round of federated
training in Gboard. The server update in Equation~\ref{eq:server_weight} is
achieved via the Momentum optimizer, using Nesterov accelerated
gradient~\cite{nesterov}, a momentum hyperparameter of 0.9, and a server
learning rate of 1.0. This technique is found to reduce training time with
respect to alternatives including pure SGD. On average, each client processes
approximately 400 example sentences during a single training epoch. The
federated CIFG converges after 3000 training rounds, over the course of which
600 million sentences are processed by 1.5 million clients. Training typically
takes 4-5 days. The top-1 recall of the federated CIFG is shown as a function of
training round in Figure~\ref{fig:accuracy_fed}. The performance of the n-gram
baseline model is also measured in the federated eval tasks to provide a
comparison for the CIFG, though the decoder is not trained in this study. N-gram
model recall is measured by comparing the decoder candidates stored in the
on-device training cache to the actual user-entered text.

\section{Results}
\label{sec:results}

The performance of each model is evaluated using the recall metric, defined as
the ratio of the number of correct predictions to the total number of tokens.
Recall for the highest-likelihood candidate is important for Gboard because
users are more prone to read and utilize predictions in the center suggestion
spot. Since Gboard includes three candidates in the suggestion strip, top-3
recall is also of interest.

\begin{table}[!h]
  \centering
  \begin{tabular}{lcc} \toprule
    Model           & Top-1 recall  & Top-3 recall  \\ \hline \midrule
    N-gram          & 13.0\%        & 22.1\%        \\
    Server CIFG     & 16.5\%        & 27.1\%        \\
    Federated CIFG  & 16.4\%        & 27.0\%        \\
  \bottomrule
  \end{tabular}
  \caption{Prediction recall for the server and federated CIFG models compared
           with the n-gram baseline, evaluated on server-hosted logs data.}
  \label{tab:eval_logs}
\end{table}

Server-hosted logs data and client device-owned caches are used to measure
prediction recall. Although each contain snippets of data from actual users, the
client caches are believed to more accurately represent the true typing data
distribution. Cache data, unlike logs, are not truncated in length and are not
restricted to keyboard usage in Google-owned apps. Thus, federated learning
enables the use of higher-quality training data in the case of Gboard.
Table~\ref{tab:eval_logs} summarizes the recall performance as measured on
server-hosted logs data, while Table~\ref{tab:eval_fed} shows the performance
evaluated with client-owned caches. The quoted errors are directly related to
the number of clients used for federated evaluation.

\begin{table}[!h]
  \centering
  \begin{tabular}{lc} \toprule
    Model           & Top-1 recall [\%]  \\ \hline \midrule
    N-gram          & $12.5 \pm 0.2$     \\
    Server CIFG     & $15.0 \pm 0.5$     \\
    Federated CIFG  & $15.8 \pm 0.3$     \\
  \bottomrule
  \end{tabular}
  \caption{Prediction recall for the server and federated CIFG models compared
           with the n-gram baseline, evaluated on client-owned data caches.}
  \label{tab:eval_fed}
\end{table}

Model performance is also measured in live production experiments with a subset
of Gboard users. Similar to top-1 recall, prediction impression recall is
measured by dividing the number of predictions that match the user-entered text
by the number of times users are shown prediction candidates. The prediction
impression recall metric is typically lower than the standard recall metric.
Zero-state prediction events (in which users open the Gboard app but do not
commit any text) increase the number of impressions but not matches.
Table~\ref{tab:eval_prod} summarizes the impression recall performance in live
experiments. The prediction click-through rate (CTR), defined as the ratio of
the number of clicks on prediction candidates to the number of proposed
prediction candidates, is also provided in Table~\ref{tab:ctr_prod}. Quoted 95\%
CI errors for all results are derived using the jackknife method with user
buckets.

\begin{table}[!h]
  \centering
  \begin{tabular}{lcc} \toprule
    Model           & Top-1 recall [\%]  & Top-3 recall [\%]  \\ \hline \midrule
    N-gram          & $5.24 \pm 0.02$    & $11.05 \pm 0.03$   \\
    Server CIFG     & $5.76 \pm 0.03$    & $13.63 \pm 0.04$   \\
    Federated CIFG  & $5.82 \pm 0.03$    & $13.75 \pm 0.03$   \\
  \bottomrule
  \end{tabular}
  \caption{Prediction impression recall for the server and federated CIFG models
           compared with the n-gram baseline, evaluated in experiments on live
           user traffic.}
  \label{tab:eval_prod}
\end{table}

\begin{table}[!h]
  \centering
  \begin{tabular}{lc} \toprule
    Model           & Prediction CTR [\%]  \\ \hline \midrule
    N-gram          & $2.13 \pm 0.03$      \\
    Server CIFG     & $2.36 \pm 0.03$      \\
    Federated CIFG  & $2.35 \pm 0.03$      \\
  \bottomrule
  \end{tabular}
  \caption{Prediction CTR for the server and federated CIFG models compared with
           the n-gram baseline, evaluated in experiments on live user traffic.}
  \label{tab:ctr_prod}
\end{table}

For both server training and federated training, the CIFG model improves the
top-1 and top-3 recall with respect to the baseline n-gram FST model. These
gains are impressive given that the n-gram model uses an order of magnitude
larger vocabulary and includes personalized components such as user history and
contacts LMs. Live user experiments show that the CIFG model also generates
predictions that are 10\% more likely to be clicked than n-gram predictions.

The results also demonstrate that the federated CIFG performs better on recall
metrics than the server-trained CIFG. Table~\ref{tab:eval_fed} shows that, when
evaluating on client cache data, the federated CIFG improves the top-1 recall by
a relative $5\%$ ($0.8\%$ absolute) with respect to the server-trained CIFG.
Comparisons on server-hosted logs data show the recall of the two models is
comparable, though the logs are not as representative of the true typing
distribution. Most importantly, Table~\ref{tab:eval_prod} shows that the
federated CIFG improves the top-1 and top-3 prediction impression recall by 1\%
relative to the server CIFG for real Gboard users. While the comparison is not
exactly apples to apples --- different flavors of SGD are used in each training
context --- the results show that federated learning provides a preferable
alternative to server-based training of neural language models.

\section{Conclusion}
\label{sec:conclusion}

We show that a CIFG language model trained from scratch using federated learning
can outperform an identical server-trained CIFG model and baseline n-gram model
on the keyboard next-word prediction task. To our knowledge, this represents one
of the first applications of federated language modeling in a commercial
setting. Federated learning offers security and privacy advantages for users by
training across a population of highly distributed computing devices while
simultaneously improving language model quality.

\section{Acknowledgements}
\label{sec:acknowledgements}

The authors would like to thank colleagues on the Google AI team for providing
the federated learning framework and for many helpful discussions.

\bibliographystyle{IEEEbib}
\bibliography{paper}

\end{document}